\documentclass[]{spie}  

\usepackage{amsmath,amsfonts,amssymb}
\usepackage{graphicx}
\usepackage{appendix}
\usepackage{array}
\usepackage[headsep=0.5cm]{geometry}
\usepackage{enumitem}
\usepackage{subfig}
\usepackage{caption}
\usepackage{url}
\usepackage{float}
\usepackage{pdfpages}
\usepackage{shortvrb}
\usepackage{mathtools}
\usepackage{multirow}
\usepackage{commath}
\usepackage{bm}
\usepackage[colorlinks=true, allcolors=blue]{hyperref}
\usepackage{subfig}
\usepackage[ruled]{algorithm2e}
\usepackage{fancyhdr}

\title{Comparison of Hand-held WEMI Target Detection Algorithms}

\author[]{Connor H. McCurley}
\author[]{James Bocinsky}
\author[]{Alina Zare}
\affil[]{Electrical and Computer Engineering, University of Florida, Gainesville, FL 32611, USA}

\authorinfo{Further author information: (Send correspondence to C.H.M)\\C.H.M: E-mail: cmccurley@ufl.edu\\  J.B.: E-mail: jbocinsky@ufl.edu \\A.Z.: E-mail: azare@ece.ufl.edu}

 \fancypagestyle{title}{%
 	\setlength{\headheight}{22pt}%
 	\fancyhf{}
 	\fancyhead[R]{\begin{tabular}{@{}l} Distribution Statement A: Approved for \\ public release. Distribution is unlimited. \end{tabular}}
 }%

\begin{document} 
\maketitle
\thispagestyle{title}


\begin{abstract}
Wide-band Electromagnetic Induction Sensors (WEMI) have been used for a number of years in subsurface detection of explosive hazards.  While WEMI sensors have proven effective at localizing objects exhibiting large magnetic responses, detecting objects lacking or containing very low amounts of conductive materials can be challenging.  In this paper, we compare a number of target detection algorithms in the literature in terms of detection performance.  In the comparison, methods are tested on two real-world data sets: one containing relatively low amounts of ground noise pollution, and the other demonstrating highly-magnetic soil interference.  Results are quantitatively evaluated through receiver-operator characteristic (ROC) curves and are used to highlight the strengths and weaknesses of the compared approaches in hand-held explosive hazard detection.
\end{abstract}


\section{INTRODUCTION}
\label{sec:introduction}  

Subsurface explosive hazards pose a significant threat to soldiers and civilians alike.  While such threats are typically buried during wartime, they can remain active for years afterward.    Timely, and effective, discovery and disposal of explosive ordinance is imperative to ensuring safe traveling conditions for troops and limiting the number of civilian causalities.  Consequently, extensive efforts have been spent in developing \textit{explosive hazard detection} (EHD) systems which are capable of uncovering this type of threat.

Subsurface object detection systems typically employ one or more sensors to extract information about an underground scene. This paper focuses on detection methods utilizing a hand-held sensor platform outfitted with \textit{wide-band electromagnetic induction} (WEMI) sensors, also called metal detectors, and a \textit{Universal Transverse Mercator} (UTM) positioning system which provides accurate spatial information.  WEMI sensors have been studied extensively for use in detecting buried, metallic objects \cite{Vijayakumar2015,Scott2007,Alvey2016,Cook2016,Dula2013}.  These sensors operate by sending a time-varying electromagnetic field through the ground via a transmit coil. Conductive subsurface objects are energized by this field and quickly return to resting states.  The emitted relaxation energy is captured as a shift in magnitude and phase across one or more of the receive coils. An array of operating frequencies is used to observe inherent responses from a wide variety of conductive objects.  While traditional WEMI approaches rely on observing high-magnitude responses to flag potential threats\cite{Fails2008}, Scott, et al. proposed that the relaxation energy emitted by conductive objects can be modeled as a function of operating frequency and used to created a dictionary of expected target responses \cite{Scott2008,Scott2010,Wei2010}.  Elaboration on the \textit{discrete spectrum of relaxation frequencies} (DSRF) target dictionary and its exploitation for use in WEMI hazard detection is provided in Section \ref{alg:dsrf}.

While there have been many successes in algorithm development for detection of high-metal targets, EHD is certainly not a solved problem.  The first obstacle making EHD difficult is that there exists a wide assortment of targets to be detected, ranging from small anti-personal to large anti-tank. Additionally, metal content can range from high to low, or even even be absent in the case of plastic targets. While it has long been assumed that the inherent properties exhibited by metal detectors make discovery of lowly and non-conductive objects difficult to achieve, recent work has shown that the utilization of a novel WEMI processing procedure may provide discriminative information for unmasking difficult targets.  Further description of this processing as well an initial investigation to the use of its generated features for hazard discovery is presented in Sections \ref{alg:projection} and \ref{sec:test3}, respectively. In addition to the range of targets, non-target objects (clutter) and natural variations in the soil's electromagnetic properties can bring rise to \textit{false alarms} (FA) where non-target objects are labeled as potential threats.  A principal challenge in EHD is mitigating false alarms while not masking true detections, or \textit{true positives} (TP).    

In this paper, the authors compare an assortment of WEMI prescreeners which have shown considerable success as subsurface object detectors on alternative WEMI hazard discovery systems.  Experiments were conducted to 1.) gauge the performance of target detectors across a variety of WEMI sensors, 2.) compare prescreeners utilizing pre-defined versus learned target concepts, and 3.) evaluate two pre-processing methods for interference removal.  Quantitative evaluation of performance is given by receiver-operating characteristic (ROC) curves which provide a measure for the probability of detection (PD) against false alarm rates (FAR).

 The remainder of this paper is organized as follows. Section \ref{sec:dataDescription} describes the hand-held WEMI data used in this paper and Section \ref{sec:methodology} elaborates on the methods explored.  Section \ref{alg:dsrf} explains the DSRF target dictionary utilized by the algorithms outlined in \ref{alg:jomp}, \ref{alg:smf} and \ref{alg:ace}. Section \ref{alg:miace}  illustrates the algorithms which learn target representatives through training. Section \ref{alg:projection} describes the novel WEMI interference removal and feature extraction proposed by Hayes, et al.  Experiment outlines and results are presented in Section \ref{sec:results} and Section \ref{sec:conclusion} concludes the paper.

\section{Description of Data}
\label{sec:dataDescription}

\subsection{Measurement Matrix}

The WEMI sensor functions by moving over regions of interest and taking measurements at discrete positions along its path. Two hand-held WEMI sensors were utilized in this study.  The sensors, hereby denoted as Sensor A and Sensor B, employ different numbers of operating frequencies to capture information at each position measurement.  It follows that the WEMI data collection can be described as a function of frequency and position.  Following the notation outlined by Hayes, et al., the WEMI measurement data can be written as $\hat{M}(\omega, p)$, where $\omega$ describes the operating frequency at which a measurement was taken, and $p$ is the corresponding spatial location at the time of measurement \cite{Hayes2017Novel,Hayes2018}. 

A measurement matrix can be created by combining a series of frequency samples through position such that
\begin{align}
	\hat{\bm{M}}(\bm{\omega}, \bm{p}) = [\hat{M}(\omega_i, p_j)]_{i,j}
\end{align}

\noindent
where $\omega_i$ is the $i^{th}$ operating frequency and $p_j$ is the $j^{th}$ position.  This combination produces a measurement matrix $\hat{\bm{M}} \in \mathfrak{C}^{K \times N}$, where $K$ frequencies were sampled at $N$ positions.

Each frequency sample collected by the sensor is stored as a complex vector representing the phasor difference between the emitted and received signals at each of the operating frequencies.  Since the complex plane $\mathfrak{C}$ is isomorphic to $\mathfrak{R}^{2}$, the real and imaginary components for each sample can be concatenated to form real-valued feature vectors 
\begin{align}
	\bm{M}(\bm{\omega}, \bm{p}) = 
	\begin{bmatrix}%
    	\mathfrak{R}(\hat{\bm{M}}(\bm{\omega}, \bm{p})) \\
    	\mathfrak{I}(\hat{\bm{M}}(\bm{\omega}, \bm{p}))
    \end{bmatrix}
\end{align}

\noindent
such that $\bm{M} \in \mathfrak{R}^{M \times N}$, with $M = 2K$,  becomes the data matrix used throughout this work.
\subsection{Measurement Model}
With the measurement matrix defined, the underlying data model can be described.  As previously delineated, the measurement matrix $\bm{M}(\bm{\omega}, \bm{p})$ can be decomposed as a superposition of four major components \cite{Hayes2018}
\begin{align}
	\bm{M}(\bm{\omega}, \bm{p}) = \bm{S}(\bm{\omega}, \bm{p}) + \bm{G}(\bm{\omega}, \bm{p}) + \bm{R}(\bm{\omega}, \bm{p}) + \mathcal{E}
\end{align}

\noindent
where $\bm{S}(\bm{\omega}, \bm{p})$ is the desired target response, $\bm{G}(\bm{\omega}, \bm{p})$ is a signal resulting from soil interference, $\bm{R}(\bm{\omega}, \bm{p})$ is the WEMI sensor's self-response, and $\mathcal{E}$ is zero-mean additive random noise following an $i.i.d.$ normal random variable $\mathcal{N}(0, \sigma^{2})$.

The majority of WEMI preprocessing procedures in the literature follow a two-stage system where the self-response component is addressed through some method of downtrack filtering before a separate processing step is utilized on the frequency information.  For now, we only address removal of the sensor self-response component as a method of data pre-processing.  Mitigation of soil-interference is addressed in Section \ref{alg:projection}.
\subsection{Down-track Filtering}
\label{alg:selfRemoval}
Self-response interference results from induced coupling between the transmit and receive coils of the WEMI sensor.  While the interference is a strong function of frequency, it can essentially be viewed as constant with respect to position.  This allows the self-response to be considered as a DC component in the downtrack dimension of the data.  The standard approach to self-response removal is to apply a high-pass downtrack filter.  A variety of filtering approaches have been investigated in the literature, including: Weiner, matched, and adaptive filtering, as well as convolution with zero-mean sine waves and application of Discrete Fourier Transform (DFT) filter banks \cite{Hayes2016MatchedFilter}. In this work, however, the method described by Scott and Hayes is adopted which uses an orthonormal \textit{discrete cosine transform} (DCT) basis matrix to remove both the DC and high-frequency spatial noise from the measurement matrix \cite{Hayes2016MatchedFilter,Hayes2017Novel}.  The DCT method operates on real cosines and extends the signal with mirroring before filtering.  These qualities are desirable as sinusoidal shapes are ideal for downtrack filtering and mirroring the signal before filtering avoids generation of  potentially harmful edge effects.  Additionally, the DCT operates using real-valued coefficients which inherently allows for faster matrix computations than complex-valued methods such as the DFT. 

To apply downtrack filtering to the measurement matrix, a projector $\bm{P_R}$ is developed as 
\begin{align}
	\bm{P_R} = \bm{\mathcal{D}}\bm{I}_{S} 
\end{align}

\noindent
where $\bm{\mathcal{D}}$ is defined as the orthonormal DCT basis matrix with $\bm{\mathcal{D}} \in \mathfrak{R}^{N \times N}$ and $\bm{I_S} \in \mathfrak{R}^{N \times N}$ is a selection matrix defined to keep the $\hat{d}$ desired cosine frequencies.  As described, the sensor self-response is mostly contained within the DC term of the spatial frequencies. High frequencies corresponding to short spatial wavelengths are expected only to contain noise.  For these reasons, $\bm{I}_S$ is chosen to keep the mid-spatial frequencies of the DCT which are believed to correspond mostly to the wavelengths of targets.  

The downtrack projection is applied to the column space of $\bm{M}$ by  
\begin{align}
	\bm{M_S} = \bm{M}\bm{P_R}
\end{align}
\noindent
so that $\bm{M_S} \in \mathfrak{R}^{M \times N}$ is the measured data matrix with the self-response and high-frequency noise removed.

After downtrack filtering has been applied, the mean of the real data components are subtracted from each sample to supply the final data matrix for prescreening.

\section{METHODOLOGY}
\label{sec:methodology}

\subsection{Discrete Spectrum of Relaxation Frequencies Dictionary}
\label{alg:dsrf}
The foundation for success in a large majority of recent WEMI target detection algorithms can be attributed to the utilization of a \textit{discrete spectrum of relaxation frequencies} (DSRF) dictionary \cite{Alvey2016,Goldberg2012}.  The DSRF is a discrete case of the \textit{Distribution of Relaxation Times} (DRT) model \cite{Wei2010}.  In contrast to the DRT which is formulated in terms of relaxation times, the DSRF utilizes relaxation frequencies to create a dictionary basis for the expected electromagnetic responses of targets.  A benefit to using the DSRF is that it is directly related to the physical properties of potential targets and is thus invariant to the relative position and orientation of the WEMI sensor \cite{Alvey2016,Cook2016,Smith2017}. The dictionary $\bm{A}$ can be created by selecting a range of possible relaxation frequency terms $\zeta$, and defining each element as
\begin{align}
	a(\omega, \zeta) = \frac{j \omega / \zeta}{1 + j \omega \zeta}
\end{align}

\noindent
It follows that the atom for the $k^{th}$ relaxation frequency can be constructed as
\begin{align}
	\bm{a}_k = 
		\begin{bmatrix}%
			\frac{j \omega_1 / \zeta_k}{1 + j \omega_1 \zeta_k} & \frac{j \omega_2 / \zeta_k}{1 + j \omega_2 \zeta_k} & \dots & \frac{j \omega_M / \zeta_k}{1 + j \omega_M \zeta_k}
		\end{bmatrix}^{T}
\end{align}

\noindent
where sensor operating frequency $\omega_m$ and the relaxation frequency $\zeta_k$ are both in radians.  In this work, $\bm{\zeta}$ was created from 100 equally log-spaced frequencies ranging from just below the minimum to just above the maximum sensor operating frequency.  A visual depiction of an 18-frequency DSRF dictionary is presented in the work by Alvey, et al. and Cook, et al. \cite{Alvey2016,Cook2016}.


When using the DSRF dictionary for target detection, it is assumed that a target sample can be represented by a linear combination of the dictionary atoms of $\bm{A}$ with a small amount of additive zero-mean Gaussian noise.  Given the fixed dictionary, algorithms can be developed to search for the expected target responses defined in the atoms of $\bm{A}$.

\subsection{Joint Orthogonal Matching Pursuits (JOMP)}
\label{alg:jomp}
One method which has adopted the DSRF for use in target detection is a well-known technique for spare signal representation, \textit{Matching Pursuits} (MP) \cite{Mallat1993,Mazhar2009}. MP is a greedy algorithm which finds linear approximations of signals by iteratively projecting them onto an over-complete dictionary. In MP, a sample $\bm{x}$ is estimated as a sparse linear combination of $p$ dictionary atoms, $\bm{D} = \{ \bm{g}_d\}^{K}_{d=1}$ where $\norm{\bm{g}_d}=1$, described by
\begin{align}
	\hat{\bm{x}} = \sum^{p-1}_{j=0}{\omega^{(x)}_{j}}{\bm{g}^{(x)}_{\bm{d}_j}}
\end{align}
\noindent
where $\hat{\bm{x}}$ represents the approximation of $\bm{x}$, ${\omega^{(x)}_j}$ is the scaling factor for the contribution of the dictionary atom ${\bm{g}^{(x)}_{\bm{d}_j}}$ to the full signal, and $ p \ll K$. MP iteratively selects dictionary elements for signal estimation by first finding the dictionary element ${\bm{g}_{\bm{d}_0}}$ which produces the largest absolute value when $\bm{x}_n$ is projected onto it
\begin{align}
	{\bm{g}^{(x)}_{\bm{d}_0}} = {\bm{g}_{\bm{d}_0}}, \quad \text{where} \ \bm{d}_0 = \underset{k}{\mathrm{argmax}} \abs{ < \bm{x}, \bm{g}_k > }
\end{align}

The residual error vector between the initial estimate $\hat{\bm{x}}$ and $\bm{x}$ is given by
\begin{align}
	{\bm{r}^{(x)}_{1}} = \bm{x} - ({\omega^{(x)}_0}{\bm{g}^{(x)}_{\bm{d}_0}})
\end{align} 
\noindent
where ${\omega^{(x)}_0} = <\bm{x},\bm{g}^{(x)}_{\bm{d}_{0}}>$ is the scaling factor for $\bm{g}^{(x)}_{\bm{d}_{0}}$.

\begin{center}
	\begin{figure}[]
		\centering
		\includegraphics[width=0.6\textwidth]{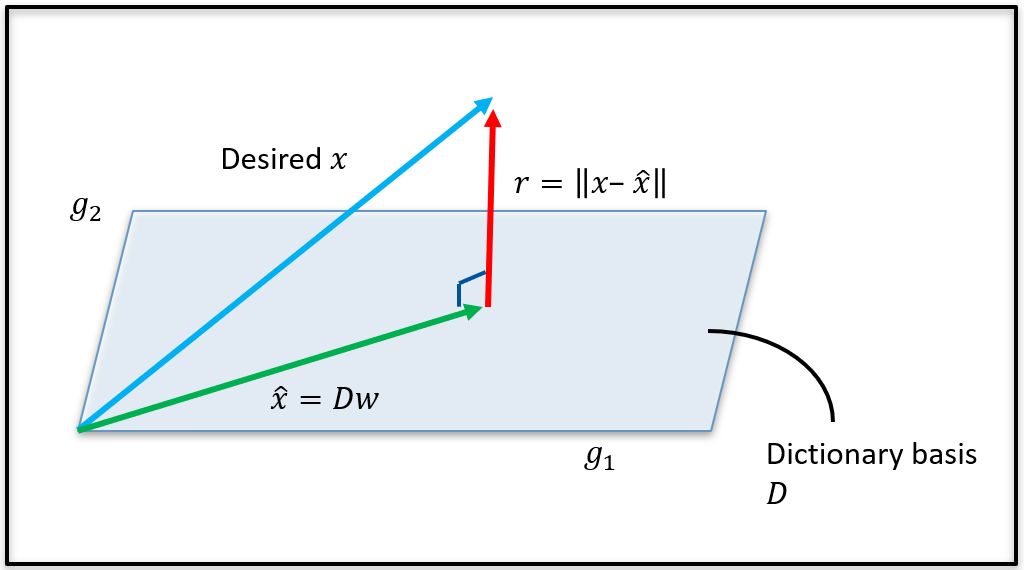}
		\caption{Geometric interpretation of MP for $\bm{x} \in \mathfrak{R}^3$ and $\bm{D}$ as a dictionary of two vectors, $g_1$ and $g_2$.  It can be observed that the residual vector is the subtraction of the projection of $\bm{x}$ onto the space spanned by $\bm{D}$.  Note that the residual vector is orthogonal to the projection onto $\bm{D}$.  }
		\label{fig:mp}
	\end{figure}
\end{center}

The signal estimation process continues iteratively by projecting the residual vectors $\bm{r}^{(x)}_i$ onto the dictionary atoms of $\bm{D}$ and updating the following residuals $\bm{r}^{(x)}_{i+1}$ for $p$ iterations.  It follows that $\bm{x}$ can be reconstructed completely by 
\begin{align}
	\bm{x} = \hat{\bm{x}} + \bm{r}^{(x)}_p = \sum^{p-1}_{j=0}{\omega^{(x)}_{j}}{\bm{g}^{(x)}_{\bm{d}_j}} + \bm{r}^{(x)}_p
\end{align}
\noindent
with ${\omega^{(x)}_{j}} = <\bm{r}^{(x)}_{j}, {\bm{g}^{(x)}_{\bm{d}_j}}>$ and $\bm{r}^{(x)}_{0} \equiv \bm{x}$.  Figure \ref{fig:mp} demonstrates the residual calculation for a three-dimensional sample reconstructed with two dictionary elements.

\vspace{-0.7cm}
An extension to MP for WEMI target detection is called \textit{Joint Orthogonal Matching Pursuits} (JOMP) \cite{Goldberg2012,Cook2016,Alvey2016, Cai2011}.  Instead of estimating a single sample signal with a given dictionary, the JOMP algorithm considers multiple signals simultaneously.  Goldberg, et al. used the DSRF dictionary and a ``two-tapped" JOMP (one sample positioned $\ell$ points ahead of the query and one sample $\ell$ behind) for explosive hazard detection.  Instead of finding the reconstruction atoms for the query sample, the dictionary elements are selected jointly between the two points.  The motivation for this method is to reduce the number of false detections attributed to single anomalous, coherent signals by only flagging signatures large enough to extend over multiple scans.  The residual error for each signal is found by $r^{(i)}_{p} = \norm{\hat{\bm{x}}^{(i)} - \bm{x}^{(i)}}^{2}_{2}$, where $i$ is the signal index. The final residual error is the average between the final residuals of the two signals $\ell$  points away from the center sample. The confidence value $c$ is assigned to the location of $\bm{x}$ at the center of the WEMI antenna by 
\begin{align}
	c(\bm{x}^{(i)}, p) = \frac{1}{1 + \frac{1}{2}(r^{(i-\ell)}_{p} + r^{(i+\ell)}_{p})}
\end{align}
\noindent
This value provides a measure of the representation ability of the DSRF for the two points being considered.  If a target is present the confidence value should be close to 1, otherwise it should be near zero.
\begin{center}
	\begin{figure}[]
		\centering
		\includegraphics[width=0.6\textwidth]{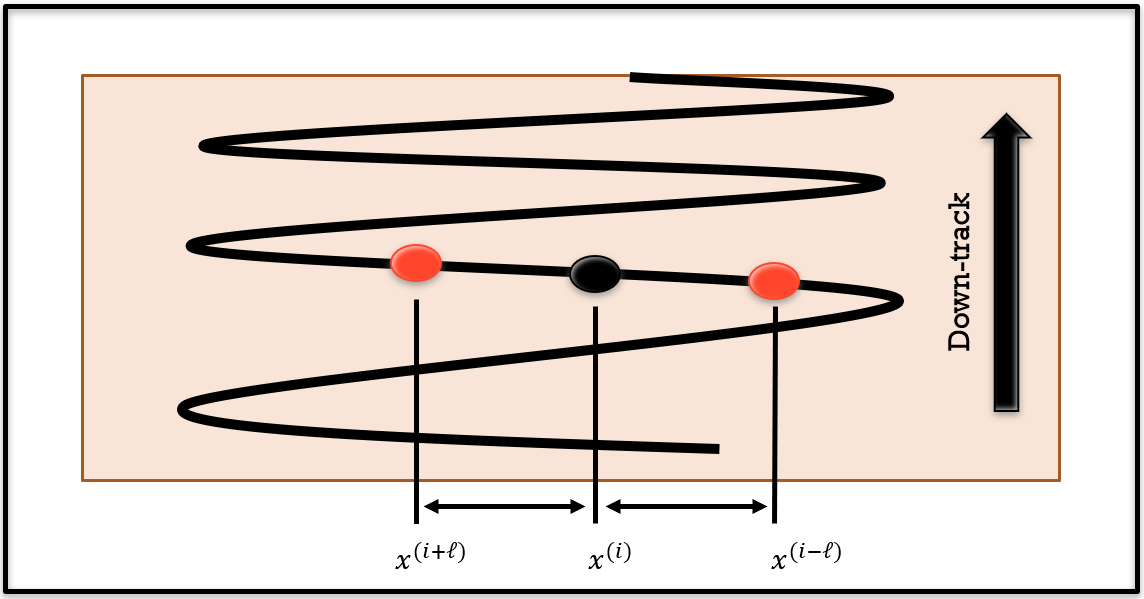}
		\caption{Example of the data used in two-tapped JOMP.  A confidence value is assigned to point $\bm{x}^{(i)}$ by jointly considering the reconstruction errors for samples $\bm{x}^{(i-\ell)}$ and $\bm{x}^{(i+\ell)}$.}
		\label{fig:jompScan}
	\end{figure}
\end{center}
\subsection{Spectral Matched Filter (SMF)}
\label{alg:smf}

The \textit{spectral matched filter} (SMF) is a statistical method for which the target and background signatures are modeled as random variables distributed according to some underlying probability distribution \cite{Zare2016}.  The detection process can be formulated as a binary hypothesis test between the null hypothesis $\mathcal{H}_0$ (target absent) and $\mathcal{H}_1$ (target present).  A detector can be modeled using the \textit{generalized likelihood ratio test}.  The hypotheses for SMF are
\begin{align}
	\begin{split}
		&\mathcal{H}_0: \bm{x} \sim \mathcal{N}(0, \bm{\Sigma}_b) \\
		&\mathcal{H}_1: \bm{x} \sim \mathcal{N}(\bm{s}, \bm{\Sigma}_b)
	\end{split}
	\label{eqt:hypothesisSMF}
\end{align}

\noindent
where $\bm{\Sigma}_b$ is the covariance of the background distribution and $\bm{s}$ is a known target signature.  The SMF detector resulting from the square root of the GLRT for Equation (\ref{eqt:hypothesisSMF}) is defined by
\begin{align}
\label{eqt:smf}
D_{SMF}(\bm{x}, \bm{s}) = \frac{\bm{s}^{T}\bm{\Sigma}^{-1}_{b}(\bm{x}-\bm{\mu}_b)}{\sqrt{\bm{s}^T\bm{\Sigma}^{-1}_{b}\bm{s}}}
\end{align}

\noindent
where $\bm{\mu}_b$ is the mean of the background which is subtracted to ensure a zero-mean distribution defined by $\mathcal{H}_0$. 

The spectral matched filter can be interpreted as a target detector which measures the inner product between a sample $\bm{x}$ and known target signature $\bm{s}$, scaled by the Mahalanobis distance of $\bm{s}$ in the background space.  This can be understood geometrically as the projection of an (unnormalized) test point onto a target vector in a whitened coordinate space.  Since test points are not normalized, data samples with larger magnitude (that are not orthogonal to the test signature) are provided larger-valued detection statistics. This quality is desirable in WEMI target detection since highly conductive buried objects will inherently induce greater magnitude responses in the sensor.

In test, the SMF statistic is computed with each element in the DSRF target dictionary.  The maximum statistic across all target representatives is assigned as the sample's final detection score.

\subsection{Adaptive Cosine Estimator (ACE)}
\label{alg:ace}

The \textit{adaptive cosine estimator} (ACE) is an alternative detection statistic which closely resembles SMF \cite{Alvey2016,Zare2016}.  The hypotheses utilized by ACE are provided by 
\begin{align}
	\begin{split}
		&\mathcal{H}_0: \bm{x} \sim \mathcal{N}(0, \sigma^{2}_{0}\bm{\Sigma}_b) \\
		&\mathcal{H}_1: \bm{x} \sim \mathcal{N}(\bm{s}, \sigma^{2}_{1}\bm{\Sigma}_b)
	\end{split}
	\label{eqt:hypothesisACE}
\end{align}
\noindent
where $\sigma^{2}_{0} = \frac{1}{M}\bm{x}^{T}\bm{\Sigma}^{-1}_b\bm{x}$ and $\sigma^{2}_{1} = \frac{1}{M}(\bm{x}-\bm{s})^{T}\bm{\Sigma}^{-1}_b(\bm{x}-\bm{s})$ to add scale-invariance with $M$ as the number of features included in the data.

The square-root of the GLRT for ACE follows as

\begin{align}
	D_{ACE}(\bm{x}, \bm{s}) = \frac{\bm{s}^{T}\bm{\Sigma}^{-1}_{b}(\bm{x}-\bm{\mu}_b)}{\sqrt{\bm{s}^T\bm{\Sigma}^{-1}_{b}\bm{s}}\sqrt{(\bm{x}-\bm{\mu}_b)^T\bm{\Sigma}^{-1}_{b}(\bm{x}-\bm{\mu}_b)}}
\end{align}

\noindent
with notation matching the SMF detector described in Section \ref{alg:smf}.  

Similar to SMF, the ACE detector can be interpreted as a measure of the inner product between a test sample and a known target signature in a whitened space.  The most prominent difference between ACE and SMF is that ACE first normalizes test samples before measuring dissimilarity.  The detection statistic then reduces simply to a measure of the vector angle between a test sample and a known target signature in a whitened space which takes values on the interval $[-1,1]$. This measure can be viewed as a confidence representing the belief that a test sample comes from the target class distribution.  Contrary to SMF, the magnitude of a test sample does not influence the ACE detection statistic value.  

As with SMF, the ACE statistic is computed with each element in the DSRF target dictionary during testing.  The maximum statistic across all target representatives is assigned as the sample's final detection confidence.

It should be noted that accurate background estimation is paramount to both ACE and SMF's success as detection statistics. Alvey, et al. explored a global background estimation and Woodbury update approach for use in ACE detection \cite{Alvey2016}.  Both methods assume that background samples follow a multivariate Gaussian random variable parameterized by a mean $\bm{\mu}_b$ and covariance $\bm{\Sigma}_b$.  In an ideal environment, background samples would all be generated from the same stationary distribution. Detection approaches known as ACE and SMF Global leverage this idea by utilizing a single training set to empirically estimate the background statistics in one pass.  While this method is non-causal, it is simple to implement and quick to execute.  The Woodbury update could be employed to allow for online training, if necessary.

\begin{center}
	\begin{figure}[]
		\centering
		\centering
		\subfloat[]{\includegraphics[width=0.4\textwidth]{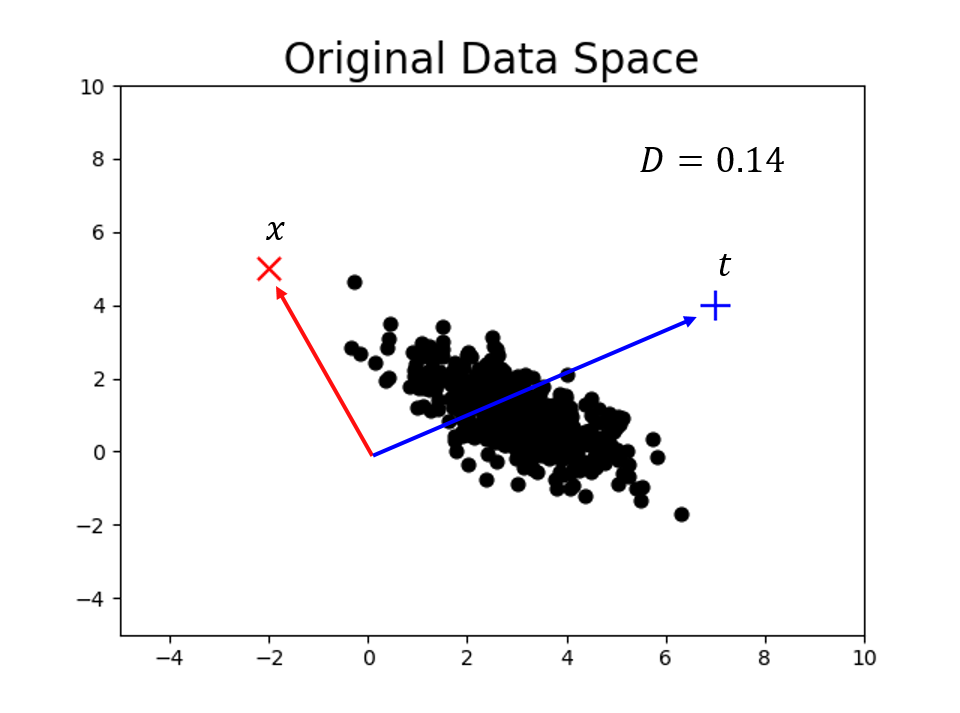}}%
		\subfloat[]{\includegraphics[width=0.4\textwidth]{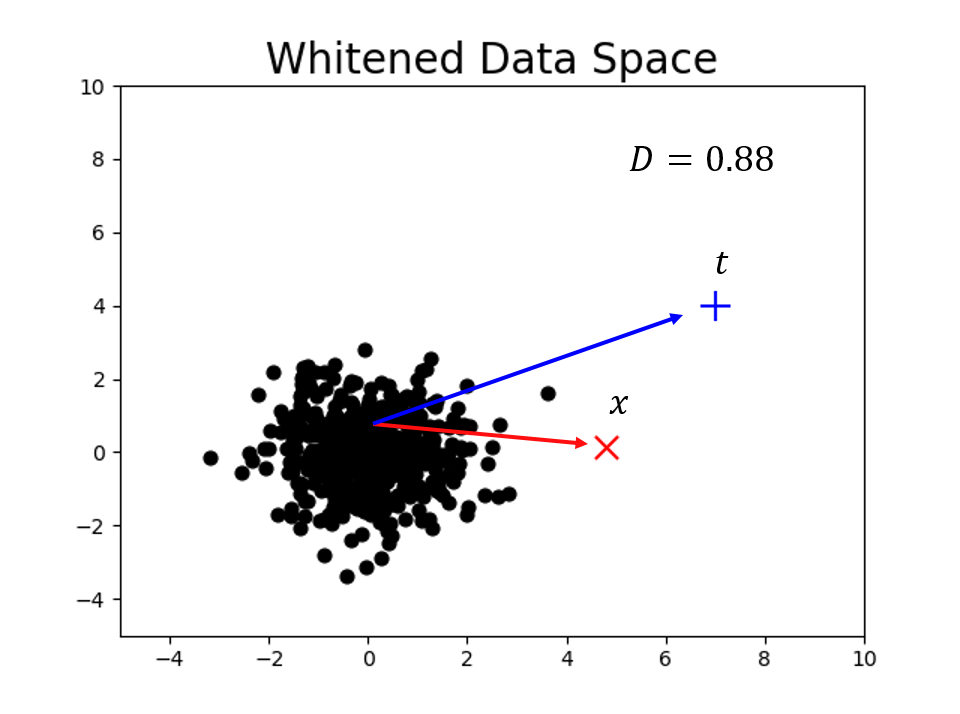}}%
		\caption{Example of target detection using both unwhitened (a) and whitened (b) data.  The cosine similarity between a known target signature, $t$, and a data sample $x$ is given as $D = 0.14$ in the original data space.  After whitening by the mean and covariance of the background samples (black) the statistic increased to $D=0.88$.     }
		\label{fig:ace}
	\end{figure}
\end{center}
\vspace{-1cm}
\subsection{Multiple Instance ACE (MI-ACE)/ SMF (MI-SMF)}
\label{alg:miace}
The WEMI hazard detection algorithms described in Sections \ref{alg:jomp}-\ref{alg:ace} utilized prior target information through employment of the DSRF dictionary.  However, it is intuitive that a target concept could be learned to take advantage of inherent properties exhibited by the data.  Traditional supervised learning tasks are difficult to implement in WEMI detection because uncertainty in the expanse of true target response often provides imprecise ground-truth. For this reason, WEMI detection tasks adhere well to \textit{multiple instance learning} (MIL) models \cite{Cook2016,Zare2016,Jiao2017,Jiao2015,Maron1998,Shrivastava2014,Zhang2001}. 

Multiple instance learning is a variation on supervised learning for problems with uncertain or imprecisely-labeled training data \cite{Zare2016}.  Instead of pairing each training sample with a class label, MIL methods learn from a set of labeled concepts called ``bags".  Each bag is defined as a multi-set of data points.  A bag is labeled as ``negative" if it is composed entirely of non-target training samples.  Alternatively, ``positive" bags are known to contain \textit{at least one} positive target sample, or \textit{instance}, although the true number of target samples within the bag is unknown.

The MIL problem is characterized by the following.  Let $\bm{X} = \begin{bmatrix} \bm{x}_1, \dots , \bm{x}_N \end{bmatrix} \in \mathfrak{R}^{M \times N}$ be training data where $M$ is the dimensionality of instance $\bm{x}_i$, and $N$ is the total number of training instances.  The training data is divided into $K$ bags, $\bm{B} = \{ \bm{B}_1, \dots , \bm{B}_K \}$ with corresponding binary bag-level labels, $L \in \{ L_1, \dots, L_K\}$ where $L_j \in \{ 0, 1\}$ and $\bm{x}_{ji} \in \bm{B}_{j}$ is the $i^{th}$ instance in bag $\bm{B}_j$.  Positive bags known to contain at least one positive target instance are given a bag-label $L_j = 1$ and are denoted by $\bm{B}^{+}_{j}$.  Alternatively, negative bags do not contain any positive target samples.  These bags are provided labels $\bm{L}_j = 0$, and are denoted $\bm{B}^{-}_{j}$.  The positive and negative bag formulations can be described succinctly by 
\begin{align}
	 L_j = 
	\begin{cases}
		1, & \exists \bm{x}_{ji} \in \bm{B}^{+}_{j} \ni \bm{x}_{ji} \sim \mathcal{N}(\bm{s} + \bm{\mu}_b, \sigma^{2}_{1} \Sigma_b) \\
		0, & \bm{x}_{ji} \sim \mathcal{N}(\bm{\mu}_b, \sigma^{2}_{0} \Sigma_b) \forall \bm{x}_{ji} \in \bm{B}^{-}_{j}
	\end{cases}
\end{align}
\noindent
with notation matching Sections \ref{alg:smf} and \ref{alg:ace}.

In contrast to ACE/ SMF Global which require an \textit{a priori} signature dictionary,  the objective of MI-ACE and MI-SMF is to learn a target signature, $\bm{s}$, that maximizes the appropriate detection statistic for the target instances in each of the positive bags while also minimizing the statistic across all negative instances.  This task is realized by maximizing the objective:
\begin{align}
	\underset{\bm{s}}{\mathrm{argmax}} \frac{1}{N^+}\sum^{}_{j:L_j = 1}D(\bm{x}^{*}_{j},\bm{s}) - \frac{1}{N^-}\sum^{}_{j:L_j = 0}\frac{1}{N^{-}_{j}} \sum^{}_{\bm{x}_i \in \bm{B}^{-}_{j}}D(\bm{x}^{*}_{i},\bm{s})
\end{align}
\noindent
where $N^+$ and $N^-$ are the number of positive and negative bags, respectively,  $N^{-}_{j}$ is the number of instances in the $j^{th}$ negative bag, and $\bm{x}^{*}_{j}$ is the instance from positive bag $\bm{B}^{+}_{j}$ which is most likely a target signature.  The most likely target signature from each positive bag is determined by
\begin{align}
	\bm{x}^{*}_{j} = \underset{\bm{x}_i \in \bm{B}^{+}_{j}}{\mathrm{argmax}} D(\bm{x}_i, \bm{s})
\end{align}
\noindent
implying that $\bm{x}^{*}_{j}$ is the training sample with the maximum detection statistic given a target signature concept, $\bm{s}$.  The training process is carried out by alternating between candidate signature selection from each of the positive bags and a statistic-dependent update step for $\bm{s}$ given the chosen candidates, $\bm{x}^{*}$'s.  Detailed derivations of the signature concept update equations for both SMF and ACE detection statistics are provided in the work by Zare, et al\cite{Zare2016}.

MI-SMF and MI-ACE have both shown considerable success in the areas of hyperspectral target and explosive hazard detection \cite{Zare2016,Alvey2017}.  This can likely be attributed to the ability of MIL methods to handle uncertainty in training.  Primary advantages of MI-ACE, as compared to alternative MIL target estimation methods, are that it does not require algorithm-specific hyper-parameters to be set and also demonstrates comparatively fast run-times.  These attributes facilitate learning of new target concepts in reliable and timely fashions. Following target concept estimation, test samples are classified in the same manner as Sections \ref{alg:smf} and \ref{alg:ace}.

\subsection{Soil Interference Removal}
\label{alg:projection}
The algorithms previously described only utilized downtrack filtering as a method for interference removal.  However, it has been shown that in some sensing environments, particular soil varieties can induce magnetic responses which are detrimental to the desired signal quality \cite{Das2006,Vijayakumar2015,Cross2008,Takahashi2013}.  An empirical soil model was constructed to define the expected signal presented by a variety of soil types \cite{Das2006,Dabas2007}.  It was determined that the soil response can mostly be characterized as a function of frequency in a two-dimensional subspace 
\begin{align}
\hat{\Psi}(\bm{\omega}) = 
\begin{bmatrix}
\bm{1} & \log{\frac{\bm{\omega}}{\omega_c}} + \frac{j \pi}{2}
\end{bmatrix}
\end{align}

\noindent
where $\hat{\Psi}(\bm{\omega}) \in \mathfrak{C}^{K \times 2}$ and $\omega_c$ is a constant valued as the geometric mean of $\bm{\omega}$.  The real and complex components can be concatenated as before to create a real-valued matrix such that $\bm{\Psi} \in \mathfrak{R}^{M \times 2}$.  

At a given position
\begin{align}
\bm{\Xi}(p) = 
\begin{bmatrix}
\xi_1(p) & \xi_2(p) 
\end{bmatrix}
\end{align}

\noindent
where $\xi_1(p)$ and $\xi_2(p)$ are the strengths of the soil response.  It follows that the soil interference can be modeled over a full sweep as 
\begin{align}
\bm{G}(\bm{\omega}, \bm{p}) = \bm{\Psi}(\bm{\omega})\bm{\Xi}^T(\bm{p})
\end{align}

\noindent
where $\bm{\Psi}(\bm{\omega}) \in \mathfrak{R}^{M \times 2}$ and $\bm{\Xi}(\bm{p}) \in \mathfrak{R}^{N \times 2}$.  It can be interpreted that the expected soil response at a given position can be decomposed as a linear combination of a DC term and a single frequency-dependent component.

While it has been assumed that WEMI sensors should only be effective at discovering targets demonstrating highly-conductive attributes, a recent work by Hayes, et al. showed that the empirical soil model may be further exploited for removing soil-response interference \cite{Hayes2017Novel, Hayes2018}.  Given appropriate conditions, this method may even be able to detect targets supplying a small conductive response or even lacking metal completely.  The method uses a projection matrix $\bm{P_G}$ to transform the data into a space which is void of soil interference.

The projection matrix $\bm{P_G}$ takes advantage of the known frequency dependencies in the measured signal $\bm{M}$.  It was shown that the soil model $\bm{G}$ can be described in terms of a two-dimensional matrix in the frequency space.  Additionally, Section \ref{alg:dsrf} demonstrated that target signals show a frequency dependence which can be modeled as a sparse collection of atoms in a DSRF dictionary, $\bm{A}$.  Hayes showed that careful combination of $\bm{A}$ and $\bm{G}$ can be exploited to project the data into a space which is mostly isolated from the soil \cite{Hayes2015SparseRecovery}.  The creation of $\bm{P_G}$ is completed in two steps.  First, a projection matrix into the soil subspace is developed as $\bm{P_{GG}} = \bm{\Psi}( \bm{\Psi}^T \bm{\Psi})^{-1} \bm{\Psi}^T$.  It follows that $\bm{A}$ can be decomposed by 
\begin{align}
\bm{A} = \bm{P_{GG}}\bm{A} + \bm{P_{GG}}^{\perp}\bm{A}
\end{align}

\noindent
with $ \bm{P_{GG}}^{\perp} = \bm{I} - \bm{P_{GG}}$ defined as the projection into the space orthogonal to the soil model.  Next, a \textit{singular value decomposition} (SVD) can be performed on the components of $\bm{A}$ such that 
\begin{align}
\bm{A} = \bm{U^{A}_{G}}{\bm{\Sigma^{A}_{G}}\bm{V^{A}_{G}}}^T + \bm{U^{A}_{\bar{G}}}\bm{\Sigma^{A}_{\bar{G}}}{\bm{V^{A}_{\bar{G}}}}^T
\end{align}

\noindent
A target SNR threshold can be enforced to allow further division between signal and noise.  In this manner, any components with a corresponding singular value below the threshold $\lambda_{\bm{G}\mathcal{E}}$ can be considered as noise.  Using this signal/ noise separation on $\bm{P_{GG}}^{\perp}\bm{A}$, the DSRF can be reformulated as
\begin{align}
\bm{A} = 
\begin{bmatrix}
\bm{U^{A}_{\bar{G}S}} & \bm{U^{A}}_{\bm{\bar{G}}\mathcal{E}} & \bm{U^{A}_{G}}
\end{bmatrix}
\begin{bmatrix}
\bm{\Sigma^{A}_{\bar{G}S}} & 0 & 0 \\
0 & \bm{\Sigma^{A}}_{\bm{\bar{G}}\mathcal{E}} & 0 \\
0 & 0 & \bm{\Sigma^{A}_{G}}
\end{bmatrix}
\begin{bmatrix}
{\bm{V^{A}_{\bar{G}S}}}^T \\
{\bm{V^{A}}_{\bm{\bar{G}}\mathcal{E}}}^T \\
{\bm{V^{A}_{G}}}^T
\end{bmatrix}
\end{align}
\noindent
which resembles an SVD but has no guarantees of orthogonality between the right singular vectors corresponding to the soil subspace and the right singular vectors representing the space orthogonal to the soil model.

This decomposition of $\bm{A}$ provides the desired orthonormal matrix for $\bm{P_G}$ as 
\begin{align}
\bm{P_G} = 
\begin{bmatrix}
\bm{P_{\bar{G}S}} \\
\bm{P}_{\bm{\bar{G}}\mathcal{E}} \\
\bm{P_{GG}}
\end{bmatrix}
\begin{bmatrix}
\bm{U^{A}_{\bar{G}S}} & \bm{U^{A}}_{\bm{\bar{G}}\mathcal{E}} & \bm{U^{A}_{G}}
\end{bmatrix}^T
\end{align}

It follows that the soil interference removal can be applied along with the self-response filtering described in Section \ref{alg:selfRemoval} by 
\begin{align}
\bm{M_S} = \bm{P_G}\bm{M}\bm{P_R}^T
\end{align}

\begin{center}
	\begin{figure}[]
		\centering
		\includegraphics[width=0.8\textwidth]{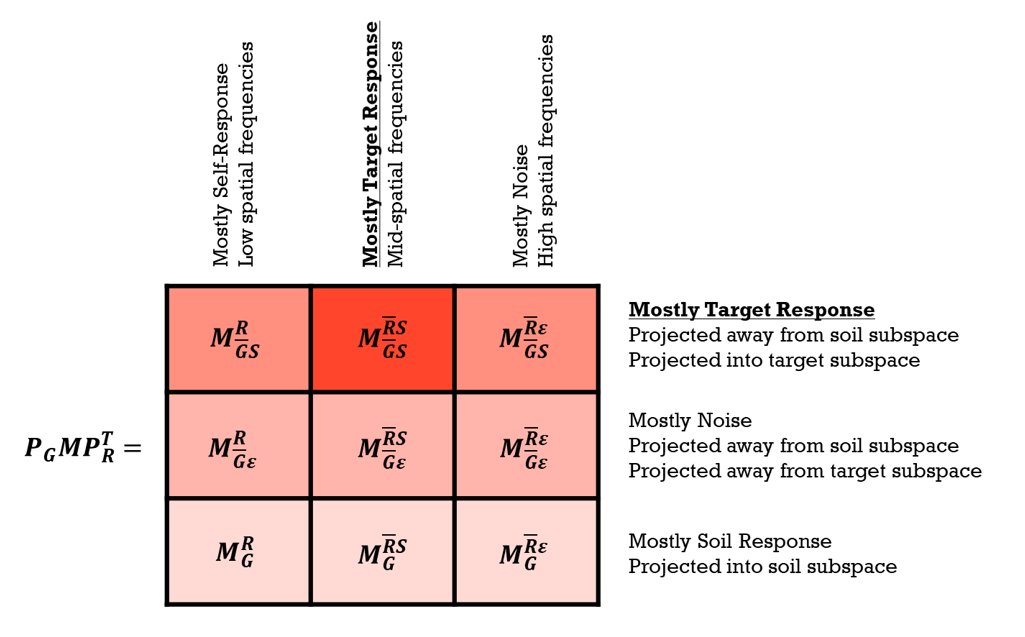}
		\caption{Data subspaces created through application of $\bm{P}_G$ and $\bm{P}_R.$ \cite{Hayes2018} }
		\label{fig:projectionBlocks}
	\end{figure}
\end{center}

As described by Hayes, application of these projections creates a data measurement matrix which can be defined in subsets demonstrating highly unique attributes \cite{Hayes2017Novel}.  This decomposition is shown in Figure \ref{fig:projectionBlocks}.

\vspace{-1cm}
Additional work by Hayes, et al. explored the value of individual sub-matrices resulting from application of the projection matrices \cite{Hayes2018}.  In this work, however, only the subspace $\bm{M}^{\bm{\bar{R}S}}_{\bm{\bar{G}S}}$ is utilized, which is believed to contain the purest portion of the signal which is void of both soil and self-response interference.  The feature generation in this work was completed by first dividing each sweep into overlapping windows centered at query samples, and applying the projection operations to each window, individually.  This process provides local information to be exploited through the SVD process.  The projection operations were performed on each data sub-matrix for the $\ell^{th}$ position of the sliding window  
\begin{align}
\bm{M}^{\bm{\bar{R}S}}_{\bm{\bar{G}S}} = \bm{P_{GS}}\bm{M}_{\ell}\bm{P_{RM}}^T  
\end{align}
\noindent
where $\bm{M}_{\ell}$ is the $\ell^{th}$ window sub-matrix of the collected data, $\bm{P_{GS}}$ is the signal portion of the soil projection and $\bm{P_{RM}}$ is the mid-spatial frequency portion of the downtrack projection matrix.

An additional SVD was performed on each data window
\begin{align}
\bm{M}^{\bm{\bar{R}S}}_{\bm{\bar{G}S}} = \bm{U_S}\bm{\Sigma_S}\bm{V_S}^T
\end{align}
and the left singular vectors $\bm{U_S}$ (corresponding to frequency) were scaled by the square root of the ratio between the corresponding and first singular values \cite{Hayes2017Novel,Hayes2018}.  The scaled representations were concatenated to create feature vectors.


\section{EXPERIMENTS AND RESULTS}{}
\label{sec:results}

\subsection{Data Set Description}
Experiments were conducted on wideband electromagnetic induction data taken from two collection sites.  Data collected from test site A has historically provided more ideal conditions for WEMI buried hazard detection.  Although the challenges detailed in Section \ref{sec:introduction} would still be present, it was anticipated that targets from this site would be detected more easily than the alternative collection location. The site disclosed as B was known to provide highly challenging conditions for WEMI hazard detection.  Expert knowledge determined that it was unlikely for low and non-metal targets to be unmasked at this location due to the strong soil-response interference emitted from this collection environment.  Test sites were divided into physical lanes where explosive hazards were buried.  Each lane was subdivided into approximately 1-2$m^2$ grids, local to each buried target.  Operators swept the WEMI sensors coarsely in sinusoidal patterns across the grids.  This is typically performed to locate potential alarms before performing more refined sweeps meant to center a discovered target.  Four operators collected at each test site over the same grids using two WEMI sensors, A and B. A UTM position coordinate was recorded with each WEMI sample.  Both data sets consist of collections across a wide assortment of explosive hazards, as well as metal and non-metal clutter.  Targets were buried at various depths and are classified by their purpose, such as anti-tank (AT) and anti-personnel (AP).  Targets are also classified by their metal content: metal target (MT), low-metal target (LMT), non-metal target (NMT), and clutter (CL).  The number of targets considered in each subset for this experimentation is shown in Table \ref{tab:target_subsets}.  Blank data believed to represent background signature void of target response was also collected at locations between grids.  Blank data was needed for both training and algorithmic scoring.

\begin{table}[t]
	\begin{minipage}{0.5\textwidth}
		\centering
			\begin{tabular}{|l | c|}
			\hline
			\multicolumn{2}{|c|}{\textbf{Number of Targets from Site A}} \\ [1ex]
			\hline
			Data Subset & Number of Targets \\ [.5ex]
			\hline
			Sensor A (Metal) & 7 \\
			\hline
			Sensor A (Low) & 10 \\
			\hline
			Sensor A (All) & 167 \\
			\hline
			Sensor B (Metal) & 9 \\
			\hline
			Sensor B (Low) & 20 \\
			\hline
			Sensor B (All) & 41 \\ [.5ex]
			\hline
		\end{tabular}
	\end{minipage}
	\begin{minipage}{0.5\textwidth}
		\centering
			\begin{tabular}{|l | c|}
			\hline
			\multicolumn{2}{|c|}{\textbf{Number of Targets from Site B}} \\ [1ex]
			\hline
			Data Subset & Number of Targets \\ [.5ex]
			\hline
			Sensor B (Metal) & 57 \\
			\hline
			Sensor B (Low) & 196 \\ [.5ex]
			\hline
		\end{tabular}
	\end{minipage}
	\caption{Number of target class instances contained in specific data subsets for test sites A and B.  Subsets are denoted by the sensor and metal content. }
\label{tab:target_subsets}
\end{table}


\subsection{Alarm Generation}
\label{subsec:alarmGeneration}
Each prescreener output a \textit{confidence map} where every UTM coordinate was associated with a value of belief that a target had been detected.  Given a confidence map, \textit{alarms} signifying points of interest were generated to mark spatial regions requiring further analysis. Alarm generation condenses groupings of high confidence samples into a single location as a way to mitigate false alarms caused by erratic, high-confidence samples.  A variety of methods for alarm generation have been explored in the literature, including: non-maximal suppression, mean-shift clustering, morphological clustering, and connected components \cite{Alvey2016,Alvey2017}.  A mean-shift algorithm with a radial basis function (RBF) kernel was used to condense the data points into alarms for scoring.  Each test sample $\bm{x}_i$, was assigned a confidence value, $c_i$.  The corresponding UTM coordinate vector is denoted by $\bm{p}_i$.  The final alarm confidence was set as the mean confidence of all points within a 0.25m radius from the converged UTM location weighted by their Euclidean distances to the center.  This weighting was performed to help mitigate spurious alarms caused by sensor cupping, centered on high-density edges of the collection.  The width of the RBF kernel was empirically determined as $\sigma = 0.075$.  The convergence threshold $\tau$ was set as $0.001$.  Pseudo-code for the mean-shift algorithm is provided in Algorithm \ref{alg:meanShift}. 
\begin{algorithm}[]
    \SetKwInput{Required}{Required}

    \Required{$\{ \bm{p}_j\}^{N}_{j=1}$, $\{ c_j\}^{N}_{j=1}$}
    \For{$j=1, \dots, N$ }{
 		$\bm{p} \leftarrow \bm{p}_j$
 		
 		\While{$\norm{\Delta \bm{p}} > \tau$}{
 			$\forall i: p(i|\bm{p}) \leftarrow \frac{c_i \exp{\left (-\frac{1}{2}\frac{\norm{\bm{p}-\bm{p}_{i}}^{2}}{\sigma}\right )}}{\sum^{N}_{i'=1}c_{i'} \exp{\left (-\frac{1}{2}\frac{\norm{\bm{p}-\bm{p}_{i'}}^{2}}{\sigma}\right )}}$

 			$\bm{p} \leftarrow \sum^{N}_{i=1}p(i|\bm{p})\bm{p}_{i}$
 		}

		$\bm{z}_j \leftarrow \bm{p}$
	}
	Connected-components$(\{ \bm{z}_j \}^{N}_{j=1}, \epsilon)$
    
    \caption{Mean Shift Alarm Generation}
    \label{alg:meanShift}
\end{algorithm}

\begin{figure}[]%
    \centering
    \subfloat[]{\includegraphics[width=0.4\textwidth]{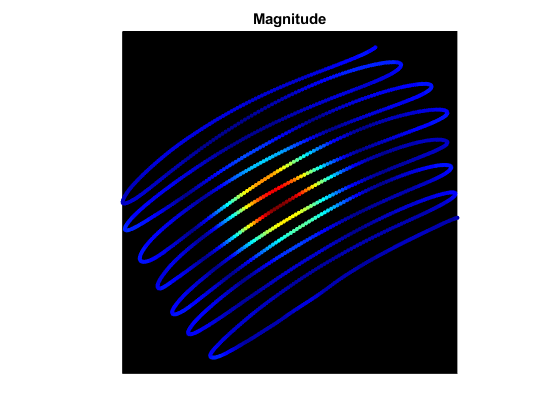}}%
    \quad
    \subfloat[]{\includegraphics[width=0.4\textwidth]{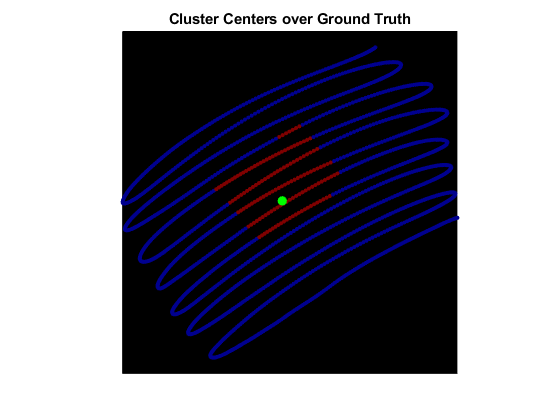}}%
    \quad \\
    \subfloat[]{\includegraphics[width=0.4\textwidth]{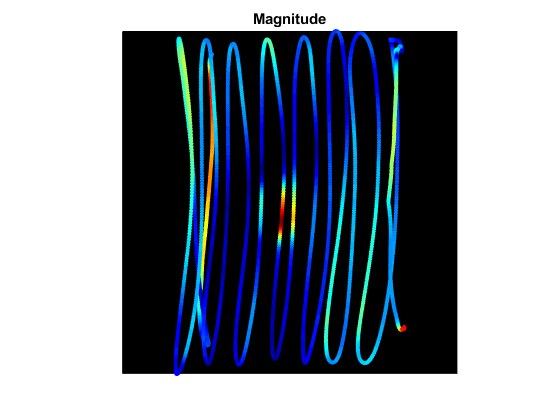}}%
    \quad
    \subfloat[]{\includegraphics[width=0.4\textwidth]{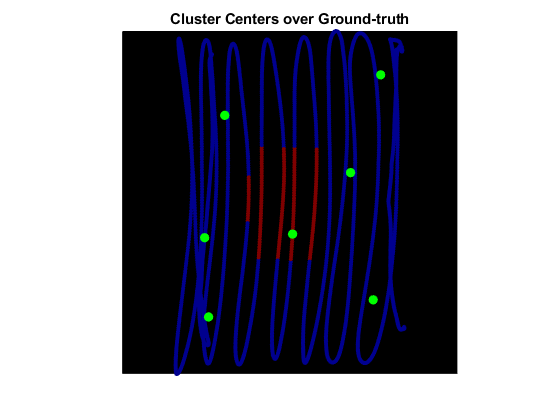}}%
    \caption{Examples of mean shift alarm generation over confidence maps. Plots (a) and (c) show confidences produced by taking the vector magnitudes of each sample at corresponding UTM locations.  Plots (b) and (d) denote the final cluster center locations determined by the mean-shift algorithm in relation to the ground-truth regions.  The mean-shift algorithm placed cluster centers at local maxima in the confidence maps. }%
    \label{fig:meanShift}
\end{figure}

\newpage
 For mean-shift, each data point begins as its own cluster center.  In each iteration of the algorithm, centers are updated towards dense regions of high confidence in the map.  The final positions of centers are located at local maxima \cite{Fukunaga1975,Carreira2015,Comaniciu2002}.  Connected components are found on the map to quantize the converged data points and the generated alarms are passed along for scoring. Examples of alarm generation using mean-shift are shown in Figure \ref{fig:meanShift}.  

\subsection{Experiment Descriptions}
In this section, the authors consider three target detection methodologies utilizing the DSRF dictionary: JOMP, ACE Global, SMF Global, as well as two methods trained and evaluated with lane-based cross-validation: MI-ACE, and MI-SMF.  A simple magnitude detector was also tested to provide a baseline.  The purpose of experimentation was three-fold.  First, comparison was made between the  WEMI sensors on Site A to determine algorithmic performance across a variety of platforms.  Next, the trained and untrained detectors were evaluated for sensor B on Site B to determine the benefits of multiple-instance learning on difficult data.  Finally, a comparison of pre-processing procedures was performed with sensor B on test site B to discover their effectiveness in highly-contaminated soil environments.  

In experimentation, each algorithm produced an alarm set signified by easting and northing UTM locations as well as confidence values.  These confidence values are not all on the same scale.  For example, JOMP is constrained to provide values between 0 and 1, ACE between -1 and 1, and SMF and magnitude greater than 0.  Since the algorithms compared are all deterministic, tests were only run once.  Mean-shift clustering was used to aggregate local regions of high-confidence in order to reduce the number of false alarms.

\subsection{Experimental Results}

\subsubsection{Comparison of Sensors}
\label{sec:test1}

\begin{center}
	\begin{figure}[]
		\centering
		\centering
		\subfloat[]{\includegraphics[width=0.33\textwidth]{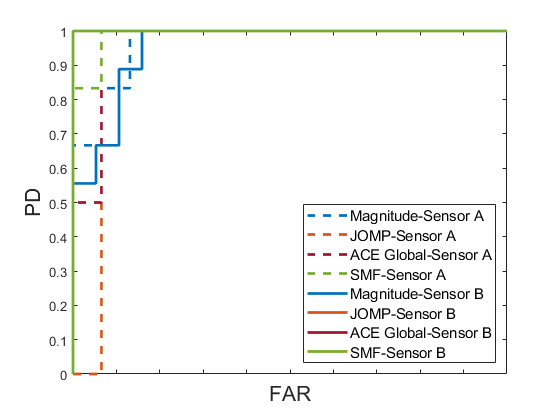}\label{fig:test1_metal}}%
		\subfloat[]{\includegraphics[width=0.33\textwidth]{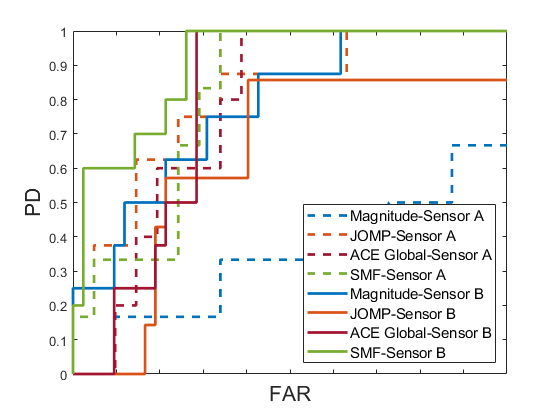}\label{fig:test1_low}}%
		\subfloat[]{\includegraphics[width=0.33\textwidth]{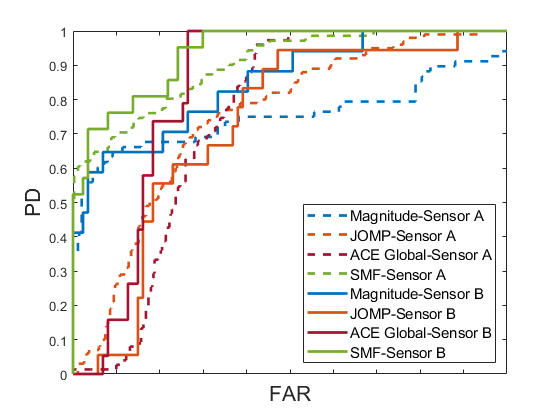}\label{fig:test1_all}}%
		\\
		\subfloat[]{\includegraphics[width=0.33\textwidth]{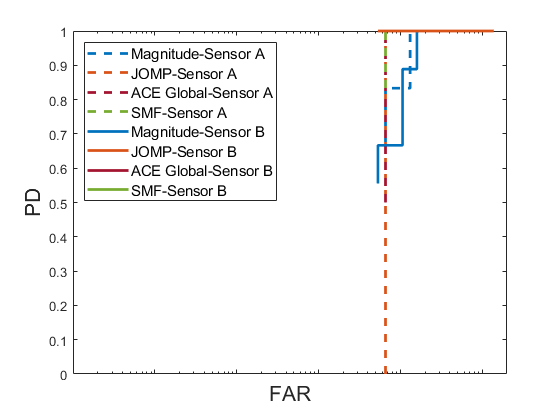}\label{fig:test1_metal_log}}%
		\subfloat[]{\includegraphics[width=0.33\textwidth]{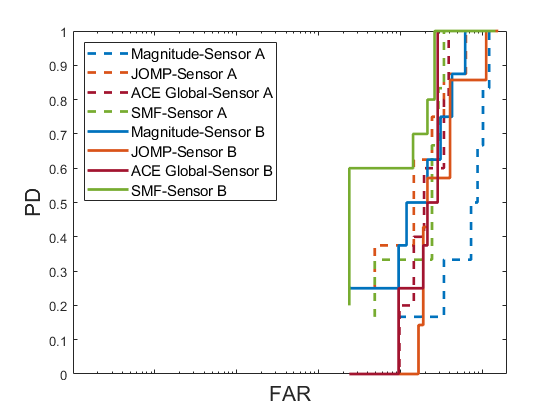}\label{fig:test1_low_log}}%
		\subfloat[]{\includegraphics[width=0.33\textwidth]{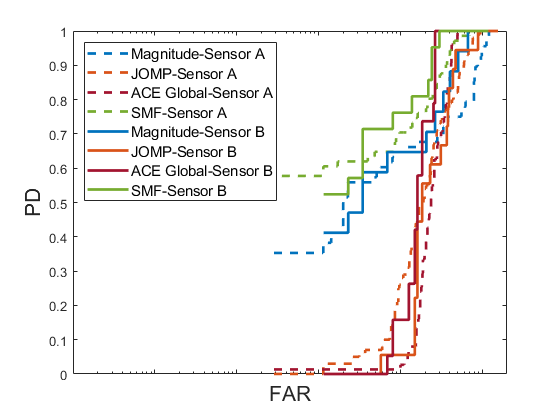}\label{fig:test1_all_log}}%
		\caption{ROC curves demonstrating the detection performance of the untrained algorithms for both sensors on test site A.  The ROCs are divided into the following subsets: \protect\subref{fig:test1_metal} high-metal,  \protect\subref{fig:test1_low} low-metal,  \protect\subref{fig:test1_all} all target types, \protect\subref{fig:test1_metal_log} high-metal plotted on a logarithmic scale,  \protect\subref{fig:test1_low_log} low-metal on a logarithmic scale, and  \protect\subref{fig:test1_all_log} all target classes on a logarithmic scale. }
		\label{fig:test1}
	\end{figure}
\end{center}

The ROC curves shown in Figure \ref{fig:test1} demonstrate the detection performance of the JOMP, ACE, SMF, and magnitude detectors for both sensors on test site A.  Data subsets included high-metal only, low-metal only, and a combination of high, low, and non-metal targets.  Figures \ref{fig:test1_metal} and \ref{fig:test1_metal_log} demonstrate that each high-metal target was discovered by every algorithm at relatively-low false alarm rates.  This result was expected as test site A had previously supplied WEMI data showing relatively low amounts of soil response interference.  Across the two sensors, SMF generally provided the best performance while the magnitude detector supplied the worst.  The performance of the magnitude detector was, again, expected.  Unlike the alternatives, the magnitude detector does not provide any mechanism for reducing false alarms.  Figures \ref{fig:test1_low} and \ref{fig:test1_low_log} show detection accuracy for low-metal targets on linear and logarithmic scales, respectively.  Figures \ref{fig:test1_all} and \ref{fig:test1_all_log} show performance on subsets containing all target classes.  As can be gleaned from the plots, SMF typically demonstrated better performance than the alternatives on both sensors A and B. The magnitude detector followed SMF, as determined by Figure \ref{fig:test1_all_log}.  While each of the detectors employing the DSRF exploit spectral shape in their decision making, only SMF considers the magnitude of a sample when assigning its confidence.  Thus, the result that SMF outperforms the alternatives implies that not only spectral shape, but also magnitude, are important features for hazard discovery.  Performance evaluation for the magnitude detector is likely less reliable for low and non-metal targets as they were not scored on the same scale as the high-metal examples.  Improved evaluation could likely be performed using a lane-based collection where data would be accumulated in a continuous stream without breaks for fiducial marking.  Each prescreener performed poorly on the test set containing high, low, and non-metal classes.  Results, however, are likely skewed due to an imbalance of target types represented in the test set. 

\vspace{-1cm}
\subsubsection{Comparison of Target Representatives}
\label{sec:test2}
\begin{center}
	\begin{figure}[H]
		\centering
		\centering
		\subfloat[]{\includegraphics[width=0.33\textwidth]{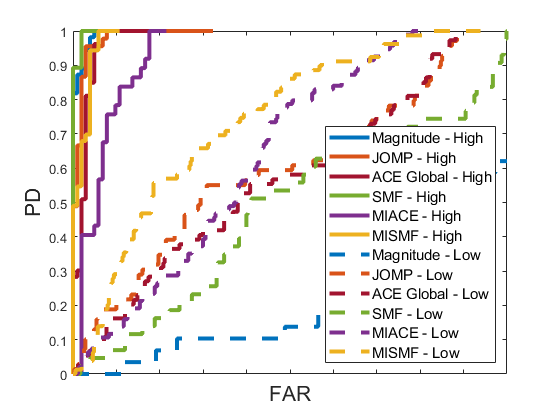}\label{fig:test2_metal_and_low}}%
		\subfloat[]{\includegraphics[width=0.33\textwidth]{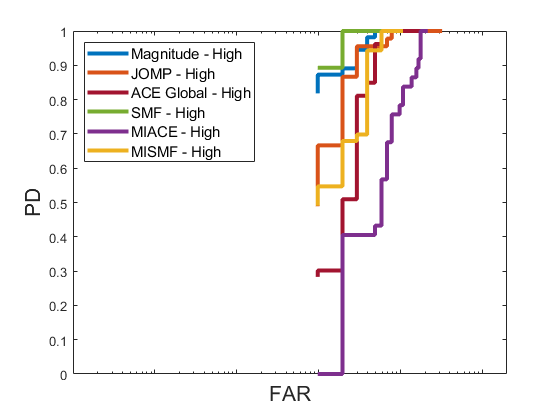}\label{fig:test2_metal_log} }%
		\subfloat[]{\includegraphics[width=0.33\textwidth]{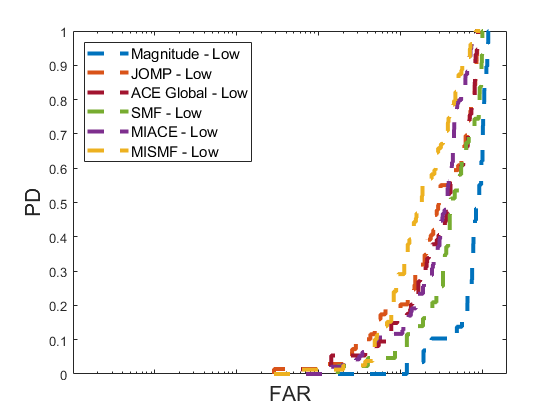}\label{fig:test2_low_log}}%
		\caption{ROC curves demonstrating the detection performance of both trained and untrained algorithms for sensor platform B on test site B.  The ROCs are divided into the following subsets: \protect\subref{fig:test2_metal_and_low} high and low-metal,  \protect\subref{fig:test2_metal_log} high-metal only on a logarithmic scale,  \protect\subref{fig:test2_low_log} low-metal only on a logarithmic scale.  Solid lines correspond to tests run on the high-metal subset, while dashed lines represent low-metal experiments.}
		\label{fig:test2}
	\end{figure}
\end{center}

A comparison of detectors utilizing the DSRF versus methods which were trained was conducted with the sensor B for test site B on high-metal and low-metal subsets.  Sensor B was chosen for this experimentation based on its frequency structure in order to facilitate comparison of preprocessing methodologies in Section \ref{sec:test3}.  Test site B was used to evaluate performance in a high-interference environment.  The ROC curves shown in Figure \ref{fig:test2} demonstrate that hazard detection was fairly less accurate in comparison with the results shown for test site A in Section \ref{sec:test1}.  This was expected due to the anticipated amounts of soil-interference presented by Site B.  Relative algorithmic performance was unchanged between test sites on the high-metal subset. In observation, prescreeners run on test Site B typically produced confidence halos with larger radii around the expected target centers than location A. This artifact could have potentially caused mean-shift to create additional alarms for the detected targets which fell outside the estimated ground-truth region, thus degrading the ROC evaluation.  This inaccuracy in scoring stems from known difficulties in providing accurate sample-level ground-truth labels for WEMI data.  Careful selection of the mean-shift kernel bandwidth was taken in attempt to mitigate this undesirable effect; although, selection of this parameter is difficult due to uncertainty in the radius of response provided by alternative targets in differing soil environments.

Because of the ground-truth uncertainty, it was hypothesized that the multiple-instance approaches taken by MI-ACE and MI-SMF would outperform the alternatives.  While this was not the case overall, multiple-instance methods showed promising results on low-metal targets.  Degradation of performance for MI-ACE and MI-SMF can most likely be attributed to the fact that training and evaluation of the algorithms was performed by learning single target representatives through lane-based cross validation.  Compared to the DSRF dictionary which was composed of 100 signatures, MI-ACE and MI-SMF each learned only a single target signature to represent all target classes.  However, this approach to training is likely sub-optimal, as experience has shown that even targets belonging to the same class can exhibit highly varying spectral signatures.  Moreover, evaluation of performance was left slightly to chance since learned target signatures had potential to be tested on specific target classes which were absent from their corresponding training folds.  Given the inherent nature of uncertainty presented in the buried hazard detection problem, MI-ACE and MI-SMF would likely outperform the alternative approaches if multiple target signatures were estimated from target-specific subsets.

Figures \ref{fig:test2_metal_and_low} and \ref{fig:test2_low_log} show that each algorithm had difficulties discovering low-metal targets, although MI-SMF performed better than the rest. This was anticipated due to the interference qualities provided by the test site.  MI-ACE and JOMP provided the next best detection performances on the low-metal subset. 

\vspace{-1cm}
\subsubsection{Comparison of Pre-processing}
\label{sec:test3}
\begin{center}
	\begin{figure}[H]
		\centering
		\includegraphics[width=0.4\textwidth]{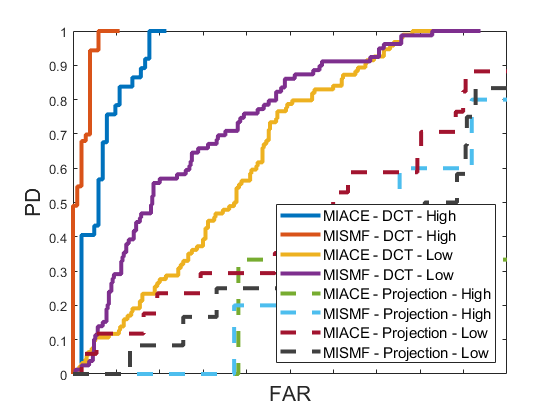}
		\caption{ ROC curves demonstrating the detection performance of  MI-ACE and MI-SMF for DCT and Projection pre-processing.  Tests were conducted on high and low-metal subsets collected with sensor platform B at test site B. Solid lines correspond to tests on data employing DCT pre-processing while dashed lines demonstrate the projection approach.}
		\label{fig:test3}
	\end{figure}
\end{center}

Following the results presented in Section \ref{sec:test2}, an investigation to the effectiveness of the soil interference removal procedure presented by Hayes, et al. was performed on the highly-conductive test environment, Site B.  In their method, a projection is applied in addition to DCT filtering in order to transform the data into a space which is orthogonal to an empirically-defined soil model.  The first three frequency singular vectors produced by this processing were weighted by ratios of their corresponding singular values and concatenated to form feature vectors.  Only MI-ACE and MI-SMF were evaluated in this test, although the soil projection could potentially be applied to the DSRF as well.  Performance results are exhibited in Figure \ref{fig:test3}.  While the same complications discussed in Section \ref{sec:test2} arose in training and evaluation, it can be observed that the DCT preprocessing  for self-response removal significantly outperformed the soil projection method for both high and low-metal subsets.  Upon further investigation, the soil projection method provided high confidence to the majority of the 1-2 $m^2$ grids under test.  The authors believe this attribute is a result of the windowing needed to perform the SVD estimation.  Because of this spatial smoothing artifact, performance of the projection processing should be evaluated on data collected in a `localization' mode where there would exist a long, continuous stream of data with well-separated targets. This type of data would be conducive to showing if there is, in fact, a difference in detection across targets,  as well as validate the authors' hypothesis that the target signature gets spread throughout the grids in the current collection.  Further testing is needed to fully evaluate the effectiveness of the soil projection method. 

\vspace{-1.5cm}
\section{CONCLUSION}
\label{sec:conclusion}
This paper compared six alternative WEMI target detection methodologies in terms of effectiveness for buried, explosive hazard discovery. As expected, high-metal targets were distinguishable by each of the algorithms compared while low and non-metal targets proved more challenging.  Four primary points were determined through experimentation.  First, it was concluded that both magnitude and spectral shape provide valuable information for exploitation in hazard discovery.  Next, results demonstrated that prescreening methods which utilize training to learn positive target concepts would likely benefit from estimating signatures for individual target types, rather than a single, global target representative.  Additionally, performance of the pre-processing designed to remove soil-interference, which did not prove efficacious in this work, is likely attributed to the the grid-based collection scheme employed at the test sites.  Investigating the use of this method in a `localization' setting where the averaging effects of windowing would be diminished would likely provide an avenue for more accurate analysis.  Finally, evaluation of WEMI target detection methods is difficult due to ground-truth uncertainty.  Development of more precise scoring metrics would greatly benefit this area of research.

\acknowledgments      
This work was funded by Army Research Office grant number W911NF-17-1-0213 to support the US Army RDECOM CERDEC NVESD. The views and conclusions contained in this document are those of the authors and should not be interpreted as representing the official policies either expressed or implied, of the Army Research Office, Army Research Laboratory, or the U.S. Government. The U.S. Government is authorized to reproduce and distribute reprints for Government purposes notwithstanding any copyright notation hereon.

\bibliography{report1} 
\bibliographystyle{spiebib} 

\end{document}